\documentclass[letterpaper]{article} 
\usepackage{aaai25}  
\usepackage{times}  
\usepackage{helvet}  
\usepackage{courier}  
\usepackage[hyphens]{url}  
\usepackage{graphicx} 
\usepackage{natbib}  
\usepackage{caption} 
\usepackage{algorithm}
\usepackage{algorithmic}

\usepackage[utf8]{inputenc} 
\usepackage[T1]{fontenc}    
\usepackage{url}            
\usepackage{booktabs}       
\usepackage{amsfonts}       
\usepackage{nicefrac}       
\usepackage{microtype}      
\usepackage{xcolor}         
\usepackage{lipsum}
\usepackage{float}
\usepackage{graphicx}
\usepackage{caption}
\usepackage{subcaption}
\usepackage{amsmath}
\usepackage{listings}
\usepackage{amssymb}
\usepackage{multirow}
\usepackage{enumitem}

\usepackage{bbold}
\usepackage{mathtools}
\usepackage{newfloat}
\usepackage{listings}

\newcommand\ours{SOLA-GCL}

\DeclareCaptionStyle{ruled}{labelfont=normalfont,labelsep=colon,strut=off} 
\floatstyle{ruled}
\newfloat{listing}{tb}{lst}{}
\floatname{listing}{Listing}
\title{\ours{}: Subgraph-Oriented Learnable Augmentation Method\\ for Graph Contrastive Learning}
\author {
    Tianhao Peng\textsuperscript{\rm 1},
    Xuhong Li\textsuperscript{\rm 2},
    Haitao Yuan\textsuperscript{\rm 3$\ast$},
    Yuchen Li\textsuperscript{\rm 2, \rm 4},
    Haoyi Xiong\textsuperscript{\rm 2}\thanks{Both Haitao Yuan and Haoyi Xiong are corresponding authors.}
}
\affiliations{
    \textsuperscript{\rm 1}Beihang University
    \textsuperscript{\rm 2}Baidu Inc.\\
    \textsuperscript{\rm 3}Nanyang Technological University
    \textsuperscript{\rm 4}Shanghai Jiao Tong University\\
    pengtianhao@buaa.edu.cn, lixuhong@baidu.com, yuchenli@sjtu.edu.cn\\ haitao.yuan@ntu.edu.sg, haoyi.xiong.fr@ieee.org
}

\usepackage{bibentry}

\begin{document}

\maketitle

\begin{abstract}
Graph contrastive learning has emerged as a powerful technique for learning graph representations that are robust and discriminative. However, traditional approaches often neglect the critical role of subgraph structures, particularly the intra-subgraph characteristics and inter-subgraph relationships, which are crucial for generating informative and diverse contrastive pairs. 
These subgraph features are crucial as they vary significantly across different graph types, such as social networks where they represent communities, and biochemical networks where they symbolize molecular interactions.
To address this issue, our work proposes a novel \textbf{S}ubgraph-\textbf{O}riented \textbf{L}earnable \textbf{A}ugmentation method
for \textbf{G}raph \textbf{C}ontrastive \textbf{L}earning, termed \ours{}, that centers around subgraphs, taking full advantage of the subgraph information for data augmentation. 
Specifically, \ours{} initially partitions a graph into multiple densely connected subgraphs based on their intrinsic properties. To preserve and enhance the unique characteristics inherent to subgraphs, a graph view generator optimizes augmentation strategies for each subgraph, thereby generating tailored views for graph contrastive learning. 
This generator uses a combination of intra-subgraph and inter-subgraph augmentation strategies, including node dropping, feature masking, intra-edge perturbation, inter-edge perturbation, and subgraph swapping.
Extensive experiments have been conducted on various graph learning applications, ranging from social networks to molecules, under semi-supervised learning, unsupervised learning, and transfer learning settings to demonstrate the superiority of our proposed approach over the state-of-the-art in GCL. 
\end{abstract}


\section{Introduction}\label{intro2}
Graph structures are used to represent data such as social networks~\cite{zhou2020graph}, molecules~\cite{DBLP:conf/icdm/0006XKSCWY23,zhang2023adaprop}, and traffic flows~\cite{yuan2021survey,yuan2023automatic}. Graph neural networks (GNNs)~\cite{DBLP:conf/iclr/KipfW17, DBLP:conf/iclr/VelickovicCCRLB18, DBLP:conf/iclr/XuHLJ19, DBLP:conf/nips/YunJKKK19} have become popular for graph representation learning~\cite{lin2023multi,tang2023diffuse,DBLP:journals/tkde/LiXWKLLBWCDY23,cai2023effective,yuan2021effective}, but they require labeled data. To learn from unlabeled data, contrastive learning has been adapted to graph data~\cite{DBLP:conf/nips/YouCSCWS20, DBLP:conf/icml/YouCSW21, DBLP:conf/iclr/VelickovicFHLBH19, DBLP:conf/nips/SureshLHN21, DBLP:conf/aaai/YinWHXZ22}. This involves generating two views by perturbing the graph and learning representations by maximizing feature consistency~\cite{DBLP:conf/nips/YouCSCWS20, DBLP:conf/icml/YouCSW21, DBLP:conf/iclr/VelickovicFHLBH19,DBLP:conf/kdd/LiXKWWCY23}. Graph-structured data are more complex than image or text data due to their properties and distribution shifts~\cite{DBLP:conf/aaai/Hassani22,DBLP:journals/tsc/LiXKZXCL23,yuan2024nuhuo}. Methods like DGI~\cite{DBLP:conf/iclr/VelickovicFHLBH19}, GRACE~\cite{zhu2020deep}, and GraphCL~\cite{DBLP:conf/nips/YouCSCWS20} use random augmentations, but lack automatic selection of augmentation policies, affecting the semantic integrity of graphs.  


Recent advances in Graph Contrastive Learning (GCL) leverage methods like JOAO~\cite{DBLP:conf/icml/YouCSW21} and LG2AR~\cite{DBLP:journals/corr/abs-2201-09830} adaptively select one augmentation strategy from a predefined pool for specific graph data, followed by random graph view generation process such as node masking and edge dropping. 
While these methods improve the adaptability of selecting augmentation strategies for individual graphs, their reliance on random view generation may change the original graph semantics.
Despite some advancements with end-to-end methods like AD-GCL and AutoGCL, which introduce learnable augmentations, the full potential of subgraph information is still largely unexplored.
Methods like MSSGCL~\cite{liu2023multi} and SUBG-CON~\cite{DBLP:conf/icdm/JiaoXZ0ZZ20} attempt to address this by focusing on the relationships between sampled subgraphs and the entire graph or the central nodes, respectively. However, they still struggle to capture the intra-subgraph characteristics and inter-subgraph relationships (e.g., communities in social networks). 
Furthermore, while these learnable methods manage to preserve the semantics of original graphs, they lack adaptability across different datasets due to their reliance on uniform augmentation strategies~\cite{DBLP:conf/icml/YouCSW21, DBLP:conf/nips/YouCSCWS20}, which ultimately limits their flexibility.
Here, we conclude the limitations of existing methods as follows.

\begin{itemize}
    \item \textit{Loss of intra-subgraph and inter-subgraph information:} 
    Graphs contain important details within subgraphs (intra-subgraph characteristics) and between different subgraphs (inter-subgraph relationships). 
    For instance, in social networks, distinct communities exhibit unique characteristics, while in chemistry, combinations of functional groups determine molecular functionalities. 
    Previous studies have not effectively captured the critical information both within and between subgraphs.
    \textit{How to effectively capture both intra-subgraph characteristics and inter-subgraph relationships} remains an ongoing challenge.
        
    \item \textit{Poor adaptability and losing semantic information:} 
    Different graph structures necessitate distinct augmentation strategies. For example, molecular graphs may require edge perturbation, whereas social community graphs might benefit more from node dropping~\cite{DBLP:conf/icml/YouCSW21, DBLP:conf/nips/YouCSCWS20}. 
    Methods like JOAO, which lack a learnable view generation process, can offer improved adaptability but at the cost of losing semantic information. 
    Conversely, methods with uniform augmentation strategies that include a learnable view generation process, such as AD-GCL, preserve semantic information but often fail to adapt effectively across diverse datasets. \textit{Improving the adaptability of GCL methods while preserving semantic integrity} remains a significant challenge.
    
\end{itemize}

To address above challenges, We propose a novel \textbf{S}ubgraph-\textbf{O}riented \textbf{L}earnable \textbf{A}ugmentation method for \textbf{G}raph \textbf{C}ontrastive \textbf{L}earning (\ours{}) to tackle existing challenges by leveraging subgraph information for data augmentation. The process begins by dividing the original graph into connected subgraphs using partitioning algorithms, such as the Louvain method~\cite{blondel2008fast} for community graphs or RDKit~\cite{landrum2013rdkit} for molecular structures, to identify functional groups. A graph view generator then applies multiple augmentation strategies (node dropping, edge perturbation, subgraph swapping) to these subgraphs, assembling them into a new graph view. This involves a subgraph augmentation selector, which learns the optimal augmentation strategy distribution; a subgraph view generator for implementing these strategies; and a subgraph view assembler. Our method demonstrated superior performance in extensive graph classification experiments across various learning tasks, outperforming state-of-the-art graph contrastive learning approaches.    

\begin{figure*}
	\centering \includegraphics[width=\linewidth]{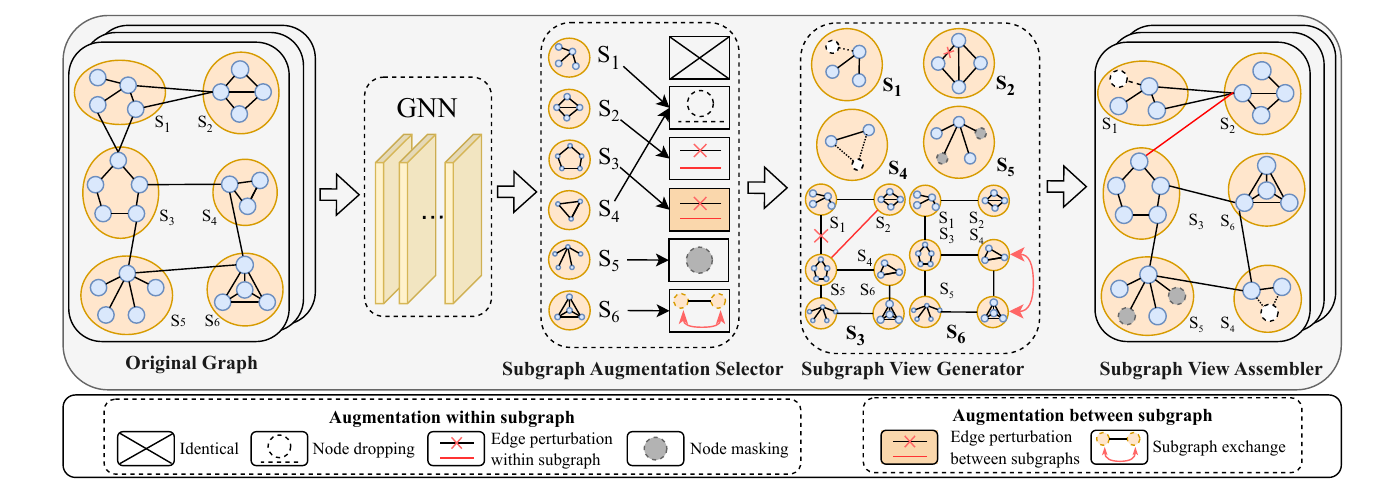} 
	\caption{An illustration of the proposed \ours{} framework. The graph view generator is composed of three critical components: the subgraph augmentation selector, the subgraph view generator, and the subgraph view assembler. The subgraph augmentation selector learns to choose the optimal augmentation strategy for each subgraph, and the subgraph view generator outputs augmented subgraph views according to the selected strategies. The subgraph view assembler constructs an augmented graph view based on these augmented subgraph views.} 
	\label{fig:framework}
\end{figure*}

The main contributions of this study are as follows: 
\begin{itemize}
    \item {A novel graph contrastive learning framework, termed \ours{}, is proposed. To the best of our knowledge, this is the first work to build learnable generative augmentation policies that specifically focus on the intra-subgraph characteristics and inter-subgraph relationships within graphs.}
    \item {The proposed \ours{} introduces an end-to-end differentiable training algorithm, enabling automatic augmentation strategy selection and graph view generation.}
    \item Extensive experiments are conducted on a variety of graph classification datasets with semi-supervised, unsupervised, and transfer learning settings, showcasing the robustness and effectiveness of our \ours{} framework on graph classification tasks.
\end{itemize}

\section{Related Work}\label{app:related}
In this section, we first introduce the studies relevant to our work from the perspectives of graph neural networks, graph partition algorithms, and graph contrastive learning (GCL) algorithms upon views generated by various graph data augmentation strategies. Later, we discuss the unique contributions made by this work compared to previous studies. 

\subsection{Graph Neural Networks}
Graph neural networks (GNNs) are the extension of the neural network models onto graph data~\cite{DBLP:journals/tnn/WuPCLZY21,DBLP:conf/pkdd/LiXKWSCCY23,zhang2023emerging,yuan2022route}. Most existing GNNs adopt the message-passing framework and use permutation-invariant local aggregation schemes to update node representations. For instance, GCN~\cite{DBLP:conf/iclr/KipfW17} averages features of all neighboring nodes. GAT~\cite{DBLP:conf/iclr/VelickovicCCRLB18} uses an attention mechanism to assign different weights to neighboring nodes. GraphSAGE~\cite{DBLP:conf/nips/HamiltonYL17} samples fixed-size neighbors of a node and aggregates their features for realizing fast and scalable GNN training. GIN~\cite{DBLP:conf/iclr/XuHLJ19} adjusts the weight of the central node by a learnable parameter to distinguish different graph structures based on the graph embedding. ResGCN~\cite{DBLP:journals/ml/PeiHIP22} combine the residual connection with GCN to build deeper GNNs. 

In this study, we employ two state-of-the-art GNNs, GIN~\cite{DBLP:conf/iclr/XuHLJ19} and ResGCN~\cite{DBLP:journals/ml/PeiHIP22}, as our backbone GNNs, following the existing graph contrastive learning literature~\cite{DBLP:conf/nips/YouCSCWS20, DBLP:conf/icml/YouCSW21, DBLP:conf/aaai/YinWHXZ22}. We believe our work could complement with the line of research on GNN while incorporating advanced GNN models for potential improvement in future studies.

\subsection{Graph Partition Algorithm}
Graph partitioning divides a graph into communities based on node and edge connectivity. Densely connected nodes form communities, while sparsely connected nodes do not. Spectral clustering approximates graph partition solutions using eigenvalue decomposition on the normalized graph Laplacian, but it is computationally expensive. The Louvain method~\cite{blondel2008fast} quickly optimizes modularity, a measure of intra-community edge density relative to inter-community edges. The Girvan–Newman (GN) algorithm~\cite{girvan2002community} identifies communities by removing edges iteratively. However, these methods often ignore node features, despite their richness in recent applications. SGCN~\cite{DBLP:journals/ijon/WangLYM21} addresses this by detecting community centers without prior labels. RDKit is a software suite for cheminformatics, computational chemistry, and predictive modeling, which offers graph partition algorithms for molecular graphs~\cite{landrum2013rdkit}. In our work, we use the Louvain method~\cite{blondel2008fast} and RDKit~\cite{landrum2013rdkit} for graph partitioning before GCL training, but our framework is adaptable to other partition algorithms, including deep learning-based approaches.    

\subsection{GCL with Data Augmentation}
In recent years, graph contrastive learning with data augmentation has attracted significant attention for self-supervised graph learning. The main idea is to maximize the agreement between representations of a graph in augmented views~\cite{li2023contre}. 

Recently, GRACE~\cite{zhu2020deep} introduces two general types of augmentations (edge perturbation and attribute masking), while GraphCL~\cite{DBLP:conf/nips/YouCSCWS20} proposes four (node dropping, edge perturbation, attribute masking, and subgraph sampling), applying these strategies randomly, which limits task adaptability. To address this, JOAO~\cite{DBLP:conf/icml/YouCSW21} and GPA~\cite{DBLP:journals/corr/abs-2209-06560} adaptively select suitable augmentation strategies by learning the distribution of graph datasets. However, these methods often apply random augmentations to specific graphs, resulting in sub-optimal performance.
To enhance adaptability in GCL models, AD-GCL~\cite{DBLP:conf/nips/SureshLHN21} uses adversarial training to perturb edges and generate graph views, while AutoGCL~\cite{DBLP:conf/aaai/YinWHXZ22} focuses on perturbing nodes for augmented views. Despite their success, these approaches often overlook the importance of subgraphs.
Methods like SUBG-CON~\cite{DBLP:conf/icdm/JiaoXZ0ZZ20} and MSSGCL~\cite{liu2023multi} have explored subgraph sampling for contrastive learning by generating global and local views at different scales. However, they typically fail to leverage the intra-subgraph characteristics and inter-subgraph relationships fully. To address these limitations, we propose a subgraph-centered method that generates augmented subgraph views and assembles them into an augmented graph, utilizing subgraph information in a learnable manner.

\subsection{Discussion on Most Relevant Works}
Our GCL framework introduces a learnable graph view generator that focuses on subgraphs to enhance graph understanding~\cite{DBLP:conf/pakdd/AdhikariZRP18, DBLP:conf/nips/0001XLW21}. The most relevant works to our study are JOAO~\cite{DBLP:conf/icml/YouCSW21} and MSSGCL~\cite{liu2023multi}, from the perspectives of model adaptability and subgraph information modeling, respectively.

From the perspective of model adaptability, both JOAO~\cite{DBLP:conf/icml/YouCSW21} and our \ours{} learn the probability distribution of data augmentation strategies. While JOAO applies the random graph view generation process that risks altering the original graph's semantics, our \ours{} generates new graph views in a learnable manner, thus preserving the semantic integrity.
From the perspective of subgraph information modeling, our proposed \ours{} method differs significantly from MSSGCL~\cite{liu2023multi}, which focuses on the relationship between the original graph and its sampled subgraph. In contrast, \ours{} method delves deeper, fully exploiting both the crucial intra-subgraph characteristics and inter-subgraph relationships by applying learnable targeted augmentation strategies within and between subgraphs.

\section{Methodology}
This section elaborates on the proposed \ours{} framework for graph classification in detail. With an overview shown in Fig.~\ref{fig:framework}, \ours{} first partitions the graph into several densely connected subgraphs, and then trains the subgraph augmentation selector and subgraph view generator jointly in an end-to-end manner to generate graph views for graph contrastive learning. 

\subsection{The Basic GNN Module}
\ours{} employs a vanilla GNN with $L$ local aggregation layers, enabling nodes to access information from $L$-hop neighbors, as the basic module. Thus, for a given node $v\in \mathcal{V}$, the $l$-th layer's calculation formula in an $L$-layer GNN ($l\text{=}1,2,...,L$) is as follows:
{\small
    \begin{flalign} 
    h_v^{(l)} &=\text{UPDATE}^{(l)} \left (h_v^{(l-1)}, \underset{\forall u\in \mathcal N(v)}{\text{AGGREGATE}}^{(l)}\left (\left\{ h_u^{(l-1)}\right\} \right) \right)
    \label{eq:gnn}
    \end{flalign} 
}where $h_v^{(l)}$ is the feature of a node $v$ in the $l$th layer, and $h_v^{(0)}\text{=}x_v$, $x_v$ is the original feature vector of node $v$; $\text{AGGREGATE}^{(l)}(\cdot)$ and $\text{UPDATE}^{(l)}(\cdot)$ represent the feature aggregation function (e.g., mean, LSTM, and max pooling) and feature update function (e.g., linear-layer combination and MLP)~\cite{DBLP:journals/sigkdd/DingXTL22}, respectively; $\mathcal N(v)$ represents the one-hop neighboring node set of a node $v$.

\subsection{Learnable Graph View Generator}
Given a graph $G\text{=}(\mathcal{V}, \mathcal{E}, X)$, where $\mathcal{V}\text{=}\{ v_1, v_2, ..., v_N\}$ denotes a node set, $\mathcal{E} \subseteq \mathcal{V} \times \mathcal{V}$ is an edge set, $X \in \mathbb{R}^{N \times d}$ represents node features, the graph view generator aims to create an augmented view without semantic labels during training. With an overview shown in Fig.~\ref{fig:framework}, the graph view generator is composed of three critical components: the \textbf{\em subgraph augmentation selector}, the \textbf{\em subgraph view generator}, and the \textbf{\em subgraph view assembler}. The subgraph augmentation selector aims to select the optimal augmentation strategy for each subgraph, then the subgraph view generator generates augmented views for subgraphs, and the subgraph view assembler generates an augmented graph view by combining the augmented subgraph views. 

\subsubsection{Subgraph Augmentation Selector}
In studying different types of networks such as social and module graphs, it becomes crucial to extract and differentiate key information specific to each network type. To effectively implement GCL on various dataset-specific subgraphs, we utilize graph partition algorithms like the Louvain algorithm~\cite{blondel2008fast}, which partitions a graph into densely connected subgraphs. For a graph $G\text{=}(\mathcal{V}, \mathcal{E}, X)$, the partition process can be formally expressed as follows: $\{\mathbf{S_1}, \mathbf{S_2}, ..., \mathbf{S_k}\} \text{=} \text{ALGO}(G)$, where $\mathbf{S_i}$ represents the $i$-th subgraph partitioned by algorithm $\text{ALGO}(\cdot)$. Later, an $L$-layer GNN processes node attributes to extract embeddings, which are then aggregated to form subgraph embeddings used for selecting appropriate augmentation strategies from a pool (node drop, feature mask, edge perturbation, subgraph swap) as detailed in~\ref{sec:subview-generator}. The Gumbel-Softmax technique~\cite{jang2016gumbelsoftmax} is employed to probabilistically assign augmentation operations to these partitions, enhancing the distinctiveness of the subgraph analysis.

For each subgraph $\mathbf{S}$, the selection of the augmentation strategy can be formulated as follows:
\begin{align}
p_\mathbf{S} &= \text{FUNC}(\text{READOUT}\left (\left\{ h_v^{(L)}:v\in  \mathbf{S} \right\} \right)) \\
f_\mathbf{S} &= \text{GumbelSoftmax}\left (p_\mathbf{S}\right)
\end{align}
where $h_v^{(L)}$ denotes the embedding of node $v$. $\text{READOUT}$ denotes a function that summarizes over all node embeddings in subgraph $S$. $\text{FUNC}$ is a function (e.g. MLP or a linear layer) that transforms the embedding of the subgraph to the probability distribution  $p_\mathbf{S}$. This distribution represents the probabilities across all potential augmentations for each subgraph.
$f_\mathbf{S}$ represents the augmentation choice for subgraph $\mathbf{S}$. $f_\mathbf{S}$ is a one-hot vector sampled from this distribution via Gumbel-Softmax and it is differentiable thanks to the reparameterization trick. 

\subsubsection{Subgraph View Generator}\label{sec:subview-generator}
For each subgraph, the subgraph view generator would select an augmentation strategy based on $f_\mathbf{S}$, and generate a subgraph view for the subgraph. In this study, we use five augmentation strategies, including three intra-subgraph strategies (node dropping, feature masking, and intra-edge perturbation) and two inter-subgraph strategies (inter-edge perturbation and subgraph swapping). The details are as follows.

(1) {\textbf{Node dropping } is conditioned on the node representations to decide which nodes within a subgraph to drop. Given the subgraph $\mathbf{S}$, the process of node dropping can be formulated as:
\begin{align}
\label{eq:node-dropping} p_v &= \text{FUNC}(h_v^{(L)}) \\
f_v &= \text{GumbelSoftmax}\left (p_v\right) \\
\tilde{X}_{drop}, \tilde{\mathcal{E}}_{drop} &= \text{AUG}_{drop}(X, \mathcal{E}, f_v, f_\mathbf{S})
\end{align} 
where $\text{FUNC}$ is a function (e.g. MLP or a linear layer) that transforms the embeddings of nodes to the probability distribution $p_v$. This distribution indicates the likelihood of each node being dropped or retained. $f_v$ is a one-hot vector sampled from this distribution via Gumbel-Softmax, $\text{AUG}_{drop}(X, \mathcal{E}, f_v, f_\mathbf{S})$ is the augmentation function that outputs the augmented node features $\tilde{X}_{drop}$ and edges $\tilde{\mathcal{E}}_{drop}$. The underlying assumption is that missing part of nodes does not damage the semantic information of the graph.
}

(2) {\textbf{Feature masking} is to mask the feature of nodes within a subgraph. The process is similar to node dropping augmentation. Feature masking implies that the absence of some node features does not affect the semantics.}

(3) {\textbf{Intra-edge perturbation} is conditioned on head and tail nodes to decide which edges that are within a subgraph to add or remove. It would be a heavy burden for back-propagation to predict the full adjacency matrix when dealing with large-scale graphs. To achieve efficient computation, we randomly sample negative edges within subgraphs. The underlying prior is that the semantic meaning of the graph is robust to the variance of edges.}

(4) {\textbf{Inter-edge perturbation} focuses on perturbing the edges between two subgraphs $\mathbf{S}_i$ and $\mathbf{S}_j$. Similarly, we first randomly sample $\left | \mathcal{E} \right |$ negative edges between subgraphs, and the process of inter-edge perturbation can be formulated as:
{\small
    \begin{align}
    \label{eq:inter-edge} p_e &= \text{FUNC}(h_v^{(L)} \parallel h_u^{(L)} \parallel h_{\mathbf{S}_i}^{(L)} \parallel h_{\mathbf{S}_j}^{(L)}: v\in \mathbf{S}_i, u\in \mathbf{S}_j) \\
    f_e &= \text{GumbelSoftmax}\left (p_e\right)\\
    \tilde{\mathcal{E}}_{inter} &= \text{AUG}_{inter}(\mathcal{E} \cup \tilde{\mathcal{E}}, f_e, f_\mathbf{S})
    \end{align} 
}where $h_{\mathbf{S}_i}^{(L)}$ is the embedding of subgraph $\mathbf{S}_i$. $\text{FUNC}$ is a function (e.g. MLP or a linear layer) that transform the embeddings of edges to the probability distribution $p_e$. This distribution indicates the likelihood of each edge that are between the subgraph $\mathbf{S}_i$ and $\mathbf{S}_j$ being dropped or retained. $\parallel$ denotes the concatenation operation, $f_e$ is a one-hot vector sampled from this distribution via Gumbel-Softmax, and $\text{AUG}_{inter}(\mathcal{E} \cup \tilde{\mathcal{E}}, f_e, f_\mathbf{S})$ is the augmentation function that outputs the augmented edge table $\tilde{\mathcal{E}}_{inter}$. The inter-edge perturbation differs from the intra-edge perturbation in that it takes into account the subgraph embedding when calculating the edge probability distribution, as shown in Eq.~\eqref{eq:inter-edge}.
}

(5) {\textbf{Subgraph swapping} is to swap the position of the subgraphs by changing the edges between subgraphs. 
The process of subgraph swapping can be formulated as:
\begin{align}
\tilde{\mathcal{E}}_{sub} &= \text{AUG}_{sub}(\mathcal{E}, f_\mathbf{S})
\end{align} 
where $\text{AUG}_{sub}(\mathcal{E}, f_\mathbf{S})$ is the augmentation function that outputs the augmented edge table $\tilde{\mathcal{E}}_{sub}$. 
It believes that most of the semantic meaning of the graph can be preserved in its local structure.}
%
The augmentation function $\text{AUG}(\cdot)$ integrates the node attribute (and adjacency matrix) with the $f_v$ (and $f_e$, the one-hot vector for edges sampled via Gumbel-Softmax) using differentiable operations such as multiplication. Consequently, the gradients of the weights of the subgraph view generator are retained in the augmented node features (edges) and can be computed using back-propagation. 

\paragraph{Subgraph View Assembler}
For a partitioned graph $\{\mathbf{S}_1, \mathbf{S}_2, ..., \mathbf{S}_k\}$, we denote the augmented subgraph view of $\mathbf{S}_i$ as $\tilde{\mathbf{S}}_i\text{=}(\tilde{X}_{\mathbf{S}_i}, \tilde{\mathcal{E}}_{\mathbf{S}_i})$. The augmented graph view $\tilde{G}\text{=}(\tilde{X}, \tilde{\mathcal{E}})$ can be obtained via $\tilde{X} \text{=} \text{ASSEMBLE}_x(\tilde{X}_{\mathbf{S}_i})$ and $\tilde{\mathcal{E}} \text{=}  \sum_{i=1}^{k}(\tilde{\mathcal{E}}_{\mathbf{S}_i})$, where $\text{ASSEMBLE}_x$ is a matrix computation operation. For the augmented graph, the edge table and node features both participate in the gradient computation, and the parameters of the graph view generator can be updated in a differentiable manner. Therefore, our graph view generator is end-to-end differentiable. 

\subsection{Complexity Analysis}

In \ours{}, we use GIN as the graph embedding model, with a complexity of $O(L \cdot (E_{\text{avg}} \cdot C_{\text{agg}} + N_{\text{avg}} \cdot C_{\text{update}}))$, where $L$ is the number of layers, $E_{\text{avg}}$ is the average number of edges, $N_{\text{avg}}$ is the average number of nodes, $C_{\text{agg}}$ is the cost of aggregation, and $C_{\text{update}}$ is the cost of updating node features. For the subgraph augmentation selector, the complexity is approximated as $O((N_{\text{avg}}+k_{\text{avg}})\cdot H)$, where $H$ is the feature dimension, and $k_{\text{avg}}$ is the average number of subgraphs. This includes an average pooling layer, a linear layer, and a Gumbel-Softmax operation. The subgraph view generator has a complexity of $O(H\cdot (N_{\text{avg}} + E_{\text{avg}}))$, involving a linear layer and a Gumbel-Softmax to generate augmented node features and edge tables. For the subgraph view assembler, the complexity is $O(k_{\text{avg}}\cdot N_{\text{avg}} \cdot H + E_{\text{avg}})$, where assembling node features has a complexity of $O(k_{\text{avg}}\cdot N_{\text{avg}} \cdot H)$ and assembling the edge table is $O(E_{\text{avg}})$, summing up the edges across all subgraphs.

In summary, the overall complexity of the framework is $O(L \cdot (E_{\text{avg}} \cdot C_{\text{agg}} + N_{\text{avg}} \cdot C_{\text{update}}) + k_{\text{avg}}\cdot N_{\text{avg}} \cdot H + E_{\text{avg}})$

\begin{table*}
    \caption{Comparison with the existing methods for unsupervised learning. \textbf{Bold} numbers indicate the best performance, while \textcolor{blue}{blue} numbers highlight the second best.}
    \centering
    \resizebox{0.75\textwidth}{!}{
        \begin{tabular}{cccccccccc}
        \hline
        Model   & MUTAG     & NCI1       & COLLAB & IMDB-B   &  IMDB-M       & Infections   & Facebook & DBLP \\
        \hline
        GL        & 81.66$\pm$2.11   & -          & -      & 65.87$\pm$0.98 & - & - & - & -   \\
        WL        & 80.72$\pm$3.00  & 80.01$\pm$0.50 & -      & 72.30$\pm$3.44 & - & - & - & - \\
        DGK       & 87.44$\pm$2.72   & {80.31$\pm$0.46} & -      & 66.96$\pm$0.56 & - & - & - & -  \\
        \hline
        node2vec  & 72.63$\pm$10.20 & 54.89$\pm$1.61 & 54.57$\pm$0.37      & 38.60$\pm$2.30   & - & - & - & -   \\
        sub2vec   & 61.05$\pm$15.80  & 52.84$\pm$1.47 & 55.26$\pm$1.54     & 55.26$\pm$1.54 & - & - & - & - \\
        graph2vec & 83.15$\pm$9.25   & 73.22$\pm$1.81 & 71.10$\pm$0.54 & 71.10$\pm$0.54 & - & - & - & -  \\
        \hline
        InfoGraph       & 89.01$\pm$1.13 & 76.20$\pm$1.06 & 70.65$\pm$1.13 & {73.03$\pm$0.87} & {49.80$\pm$1.50} & \textcolor{blue}{53.50$\pm$1.38} & {52.35$\pm$1.12} & \textcolor{blue}{50.72$\pm$1.51} \\
        GraphCL  & 86.80$\pm$1.34 & 77.87$\pm$0.41 & {71.36$\pm$1.15} & 71.14$\pm$0.44 & {49.20$\pm$1.62} & {50.50$\pm$1.04} & {49.05$\pm$0.69} & {48.81$\pm$1.10}\\
        JOAO & 87.35$\pm$1.02 & 78.07$\pm$0.47 & 69.50$\pm$0.36 & 70.21$\pm$3.08 & - & - & - & - \\
        JOAOv2  & 87.67$\pm$0.79 & 72.99$\pm$0.75 & 70.40$\pm$2.21 & 71.60$\pm$0.86 & {48.73$\pm$1.54} & {47.00$\pm$2.09} & {48.54$\pm$1.01} & {50.35$\pm$1.77} \\
        AD-GCL  & -  & 69.67$\pm$0.51 & 73.32$\pm$0.61 & 71.57$\pm$1.01 & \textcolor{blue}{49.87$\pm$1.44} & {52.50$\pm$1.08} & \textcolor{blue}{52.66$\pm$1.67} & {48.21$\pm$1.93} \\
        AutoGCL & {88.64$\pm$1.08}  & \textcolor{blue}{82.00$\pm$0.29} & 70.12$\pm$0.68 & \textcolor{blue}{73.30$\pm$0.40} & {48.47$\pm$1.88} & {49.00$\pm$1.43} & {49.83$\pm$1.50} & {49.40$\pm$1.74} \\
        SimGRACE & 89.01$\pm$1.31 & 79.12$\pm$0.44 & 71.72$\pm$0.82 & 71.30$\pm$0.77 & - & - & - & - \\
        MSSGCL & \textcolor{blue}{89.68$\pm$0.57} & 81.45$\pm$0.48 & \textcolor{blue}{73.48$\pm$0.83} & 73.14$\pm$0.38 & - & - & - & - \\
        
        \textbf{Ours} & \textbf{90.49$\pm$2.22} &  \textbf{82.07$\pm$1.32} & \textbf{73.80$\pm$1.32} & \textbf{74.20$\pm$1.89} & \textbf{51.27$\pm$1.05} & \textbf{53.50$\pm$1.92} & \textbf{52.97$\pm$1.42} & \textbf{53.25$\pm$1.90}\\
        \hline
        \end{tabular}
    }
\label{tab:unsup-exp}
\end{table*}    

\begin{table*}
    \caption{Comparison with the existing methods for transfer learning. }
        \centering
    \resizebox{0.6\textwidth}{!}{
        \begin{tabular}{lccccccccc}
        \hline
        Model                 & BBBP            & ToxCast    & SIDER      & ClinTox    & HIV        & BACE     \\
        \hline
        \emph{No  Pretrain}     & 65.8$\pm$4.5     & 63.4$\pm$0.6   & 57.3$\pm$1.6   & 58.0$\pm$4.4    & 75.3$\pm$1.9   & 70.1$\pm$5.4  \\ \hline
        Infomax                 & 68.8$\pm$0.8     & 62.7$\pm$0.4   & 58.4$\pm$0.8   & 69.9$\pm$3.0   & 76.0$\pm$0.7   & 75.9$\pm$1.6  \\
        EdgePred                & 67.3$\pm$2.4    & {64.1$\pm$0.6}   & 60.4$\pm$0.7   & 64.1$\pm$3.7   & 76.3$\pm$1.0   & {79.9$\pm$0.9}   \\
        AttrMasking             & 64.3$\pm$2.8   & \textcolor{blue}{64.2$\pm$0.5}   & 61.0$\pm$0.7   & 71.8$\pm$4.1    & 77.2$\pm$1.1   & 79.3$\pm$1.6   \\
        ContextPred             & 68.0$\pm$2.0   & 63.9$\pm$0.6   & 60.9$\pm$0.6   & 65.9$\pm$3.8     & 77.3$\pm$1.0   & 79.6$\pm$1.2  \\
        GraphCL                 & 69.68$\pm$0.67 & 62.40$\pm$0.57 & 60.53$\pm$0.88 & 75.99$\pm$2.65  & \textcolor{blue}{78.47$\pm$1.22} & 75.38$\pm$1.44 \\
        JOAOv2                 & {71.39$\pm$0.92}  & 63.16$\pm$0.45 & 60.49$\pm$0.74 & 80.97$\pm$1.64  & 77.51$\pm$1.17 & 75.49$\pm$1.27 \\
        AD-GCL                 & 70.01$\pm$1.07  & 63.07$\pm$0.72 & \textbf{63.28$\pm$0.79} & {79.78$\pm$3.52} & 78.28$\pm$0.97 & 78.51$\pm$0.80\\
        AutoGCL & \textcolor{blue}{73.36$\pm$0.77} & 63.47$\pm$0.38 & {62.51$\pm$0.63} & \textcolor{blue}{80.99$\pm$3.38}  & {78.35$\pm$0.64} & \textbf{83.26$\pm$1.13} \\
        \textbf{Ours} & \textbf{74.08$\pm$0.82} & \textbf{64.50$\pm$0.41} & \textcolor{blue}{62.79$\pm$0.91} & \textbf{81.49$\pm$1.31} & \textbf{78.68$\pm$1.01} & \textcolor{blue}{82.64$\pm$1.01} \\
        \hline
        \end{tabular}
    }
    \label{tab:transfer-exp}
\end{table*}

\subsection{Graph Contrastive Learning in \ours{}}
\label{sec-gcl}
In this work, we define contrastive loss $\mathcal{L}_{\text{cl}}$, similarity loss $\mathcal{L}_{\text{sim}}$, and classification loss $\mathcal{L}_{\text{cls}}$. The contrastive loss enforces maximizing the consistency between positive pairs $z_i, z_j$ compared with negative pairs. The similarity loss minimizes the mutual information between the views generated by the two view generators. The classification loss is used in the semi-supervised learning task to encourage the graph view generator to generate label-preserving augmentations.

For the contrastive loss, we follow the previous works~\cite{DBLP:conf/icml/ChenK0H20,DBLP:conf/nips/YouCSCWS20, DBLP:conf/aaai/YinWHXZ22} and use the normalized temperature-scaled cross entropy loss (NT-XEnt)~\cite{DBLP:conf/nips/Sohn16}. The cosine similarity function is defined as $\text{sim}(\boldsymbol{z}_i, \boldsymbol{z}_j)\text{=}\frac{\boldsymbol{z}_i\cdot \boldsymbol{z}_j}{{\lVert \boldsymbol{z}_i \rVert}_2 \cdot {\lVert \boldsymbol{z}_j \rVert}_2}$. During the training process, a data batch of $M$ graphs is randomly sampled and we pass the batch to the two graph view generators to obtain $2M$ graph views. The two augmented views from the same input graph are regarded as the positive view pair. We denote $\ell_{(i,j)}$ as the instance-level contrastive loss between a positive pair of samples $(i,j)$, and the contrastive loss of this data batch $\mathcal{L}_{\text{cl}}$ can be formulated as:
\begin{align}
\ell_{(i,j)} &= - \log \frac{\exp ( \text{sim} (\boldsymbol{z}_i, \boldsymbol{z}_j) / \tau )}{ \sum_{k=1, k\neq i}^{2M} \exp(\text{sim} (\boldsymbol{z}_i, \boldsymbol{z}_k) / \tau ) } \\
\mathcal{L}_{\text{cl}} &= \frac{1}{2M} \sum_{k=1}^{N} [\ell(2k-1, 2k) + \ell(2k, 2k-1)]
\end{align}
where $\tau$ is the temperature parameter. The final loss is computed across all positive pairs per batch.

For the similarity loss, we simultaneously minimize the mutual information between the augmentation selections generated by the two subgraph augmentation selectors and between the views generated by the two subgraph view generators. During the view generation process, the subgraph augmentation selector outputs a sampled state matrix $S$ indicating the corresponding augmentation operation of each subgraph, and the subgraph view assembler outputs the edge table $\overline{\mathcal{E}}$. As the augmentation process involves edge negative sampling, the length of edge table $\overline{\mathcal{E}}$ is not equal between the two view generators. We transform the edge table $\overline{\mathcal{E}}$ to adjacency matrix $A$ to compute the similarity loss between views. For a graph $G$, we denote the sampled state matrix of each view generator as $S_1, S_2$, and the adjacency matrix as $A_1, A_2$. The similarity loss can be formulated as:

\begin{align}
\mathcal{L}_{\text{sim}} &= \text{sim}(S_1,S_2) + \text{sim}(A_1,A_2)
\end{align}

For the classification loss, we employ the cross entropy loss $\ell_{cls}$. Give a graph sample $G$ with its corresponding class label $y$, and its augmented views denoted as $\tilde{G_1}$ and $\tilde{G_2}$, with $\text{CLS}$ representing the classifier, the classification loss $\ell_{cls}$ is expressed as follows:
{\small
    \begin{align}
    \mathcal{L}_{\text{cls}} &= \ell_{cls}(\text{CLS}(G), y) + \ell_{cls}(\text{CLS}(\tilde{G_1}), y) + \ell_{cls}(\text{CLS}(\tilde{G_2}), y)
    \end{align}
}
$\mathcal{L}_{\text{cls}}$ is utilized in the pre-training phase of the semi-supervised learning task to encourage the view generator to generate label-preserving augmentations.

\section{Experiments}
In this section, we assess the performance of \ours{} on graph classification across various applications by comparing it against state-of-the-art (SOTA) methods in semi-supervised, unsupervised, and transfer learning tasks. We also explore the effectiveness of \ours{} through visualizations of subgraph importance. Additionally, we conduct an ablation study to determine the impact of different components and provide analyses on running time to demonstrate its efficiency.  

\subsection{Experimental Settings}
We compare \ours{} with the SOTA methods on 16 datasets under unsupervised, semi-supervised and transfer learning settings. The baseline models are as follows: 
(1) three SOTA kernel-based methods: graphlet kernel (GL)~\cite{shervashidze2009efficient}, Weisfeiler-Lehman sub-tree kernel (WL)~\cite{shervashidze2011weisfeiler}, and deep graph kernel (DGK)~\cite{yanardag2015deep};
(2) six unsupervised graph-level representation learning methods: node2vec~\cite{grover2016node2vec}, sub2vec~\cite{adhikari2018sub2vec}, graph2vec~\cite{narayanan2017graph2vec}, EdgePred~\cite{hu2019strategies}, AttrMasking~\cite{hu2019strategies}, and ContextPred~\cite{hu2019strategies}; 
(3) six classic GCL methods that randomly generate graph views: InfoGraph~\cite{suninfograph}, Infomax~\cite{velivckovic2018deep}, GCA~\cite{zhu2021graph}, GraphCL~\cite{DBLP:conf/nips/YouCSCWS20}, JOAO~\cite{DBLP:conf/icml/YouCSW21}, and JOAOv2~\cite{DBLP:conf/icml/YouCSW21};
(4) four learning-based GCL method: AD-GCL~\cite{DBLP:conf/nips/SureshLHN21}, AutoGCL~\cite{DBLP:conf/aaai/YinWHXZ22}, SimGRACE~\cite{xia2022simgrace}, and MSSGCL~\cite{liu2023multi}.
The detailed description of datasets and experimental settings for unsupervised, semi-supervised, and transfer learning are shown in the appendix.

\subsection{Unsupervised Representation Learning}
As shown in Table~\ref{tab:unsup-exp}, several observations can be made:
(1) The consistent top performance across diverse datasets suggests that our method is not only versatile but also highly effective in extracting meaningful patterns from complex graph structures, surpassing both traditional graph learning methods and SOTA deep learning approaches. 
(2) Traditional methods like WL and DGK focus on graph kernels and rely on explicit feature engineering. They lack the adaptability of deep learning models, which can automatically extract features through training. 
(3) Deep learning methods like AD-GCL and AutoGCL show varying performance across datasets, illustrating the nuanced capabilities and limitations of these approaches.
(4) MSSGCL can potentially capture nuanced relationships within the graph data by focusing on subgraph and graph features. 
In contrast, our approach leverages a wider spectrum of subgraph interactions, not merely focusing on local versus global relationships but also extensively exploring the intricate relationships within and between subgraphs. The superior performance across various datasets demonstrate the effectiveness of our \ours{}.

\subsection{Transfer Learning}
According to results shown in Table~\ref{tab:transfer-exp}, three observations can be made:
(1) Our model consistently achieves SOTA or second-best performance across all evaluated datasets, demonstrating its robustness and effectiveness under the transfer learning setting.
(2) In the SIDER and BACE datasets, our model trails behind AD-GCL and AutoGCL, likely due to their more effective handling of sparse and imbalanced data typical in side effect prediction and biochemical field.
(3) AD-GCL and AutoGCL often outperform other models like GraphCL and JOAO due to their advanced strategies in a learnable manner. 
In contrast, GraphCL and JOAO typically employ more generalized augmentation and optimization strategies, which may not effectively capture the nuanced details of the data during large-scale pre-training

\subsection{Semi-Supervised Learning}
For semi-supervised learning, we perform the semi-supervised graph classification experiments on TUDataset~\cite{DBLP:journals/corr/abs-2007-08663}. As shown in Table~\ref{tab:semi-exp}, three observations can be made:
(1) Our \ours{} consistently achieves SOTA or comparative performance compared to other models due to its effective use of subgraph-specific information, enhancing learning from limited labeled data across various datasets.
(2) MSSGCL ranks second on PROTEINS and DD because of its effective multi-scale subgraph sampling strategy. However, it falls short of \ours{} which better captures and utilizes subgraph interactions.

\begin{table}
    \caption{Comparison with existing methods and different strategies for semi-supervised learning. }
    \centering
    \resizebox{0.43\textwidth}{!}{
        \begin{tabular}{lcccccccc}
        \hline
        Model                   & PROTEINS   & DD         & NCI1       & COLLAB     & IMDB-B  \\
        \hline
        \emph{Full Data}        & 78.25$\pm$1.61 & 80.73$\pm$3.78 & 83.65$\pm$1.16 & 83.44$\pm$0.77 & 76.60$\pm$4.20  \\ 
        \hline
        10\% Data               & 69.72$\pm$6.71 & 74.36$\pm$5.86 & 75.16$\pm$2.07 & 74.34$\pm$2.00 & 64.80$\pm$4.92 \\
        10\% GAE & 70.51$\pm$0.17 & 74.54$\pm$0.68 & 74.36$\pm$0.24 & 75.09$\pm$0.19 & - \\
        10\% Infomax & 72.27$\pm$0.40 & 75.78$\pm$0.34 & 74.86$\pm$0.26 & 73.76$\pm$0.29 & - \\
        10\% ContextPred & 70.23$\pm$0.63 & 74.66$\pm$0.51 & 73.00$\pm$0.30 & 73.69$\pm$0.37 & - \\
        10\%   GCA              & 73.85$\pm$5.56 & 76.74$\pm$4.09 & 68.73$\pm$2.36 & 74.32$\pm$2.30 & \textcolor{blue}{73.70$\pm$4.88}  \\
        10\%   GraphCL      & 74.21$\pm$4.50 & 76.65$\pm$5.12 & 73.16$\pm$2.90 & 75.50$\pm$2.15  & 68.10$\pm$5.15 \\
        10\% JOAO & 72.13$\pm$0.92 & 75.69$\pm$0.67 & 74.48$\pm$0.27 & 75.30$\pm$0.32 & - \\
        10\%   JOAOv2           & 73.31$\pm$0.48 & 75.81$\pm$0.73 & 74.86$\pm$0.39 & 75.53$\pm$0.18 & - \\
        10\%   AD-GCL           & 73.96$\pm$0.47 & 77.91$\pm$0.73 & \textbf{75.18$\pm$0.31} & 75.82$\pm$0.26 & - \\
        10\%   AutoGCL    & 75.65$\pm$2.40 & {77.50$\pm$4.41} & {73.75$\pm$2.25} & \textbf{77.16$\pm$1.48} & {71.90$\pm$4.79} \\
        10\% SimGRACE & 74.03$\pm$0.51 & 76.48$\pm$0.52 & 74.60$\pm$0.41 & 74.74$\pm$0.28 & - \\
        10\% MSSGCL & \textcolor{blue}{75.76$\pm$0.52} & \textcolor{blue}{78.89$\pm$0.18} & 74.77$\pm$0.31 & 76.02$\pm$0.13 & - \\
        
        10\%   \textbf{Ours}    & \textbf{77.63$\pm$3.10} & \textbf{79.43$\pm$4.31} & \textcolor{blue}{74.94$\pm$1.96} & \textcolor{blue}{76.97$\pm$1.31} & \textbf{74.50$\pm$4.65}  \\
        \hline
        \end{tabular}
        }
    \label{tab:semi-exp}
    
\end{table}

\begin{table}
\caption{Ablation study on subgraph partition, learnable subgraph augmentation selector, and learnable subgraph view generator.}
\centering
\resizebox{0.4\textwidth}{!}{
    \begin{tabular}{ccccccccc}
    \hline
       & MUTAG  & NCI1       & COLLAB & IMDB-B     \\
    \hline
       \#Avg. Nodes & 17.93  & 29.87 & 74.49 & 19.77 \\
       \#Avg. Subgraph (Louvain) & 3.94 & 4.91 & 3.04 & 2.75 \\
       \#Avg. Subgraph (GN) & 2.00 & 2.18 & 2.00 & 2.00 \\
    \hline
    \ours{} (Louvain) & {90.49$\pm$2.22} & {82.07$\pm$1.32} & {73.80$\pm$1.32} & {74.20$\pm$1.89}  \\
    
    \ours{} (GN) & {89.88$\pm$2.98} & {81.56$\pm$2.11} & {72.32$\pm$1.88} & {74.10$\pm$2.84} \\
    
    \ours{}-P & {88.33$\pm$2.04} & {79.93$\pm$1.74} & {70.92$\pm$1.91} & {71.70$\pm$2.80} \\
    
    \ours{}-S & {88.80$\pm$3.46} & {81.58$\pm$1.57} & {71.86$\pm$1.63} & {74.00$\pm$2.63} \\
    
    \ours{}-R & {88.80$\pm$3.24} & {80.95$\pm$1.75} & {69.38$\pm$1.96} & {73.70$\pm$2.84} \\
    \hline
    \end{tabular}
}
\label{tab:ablation-exp}
\end{table}    

\subsection{Visualization}
\begin{figure}
	\centering \includegraphics[width=0.9\linewidth]{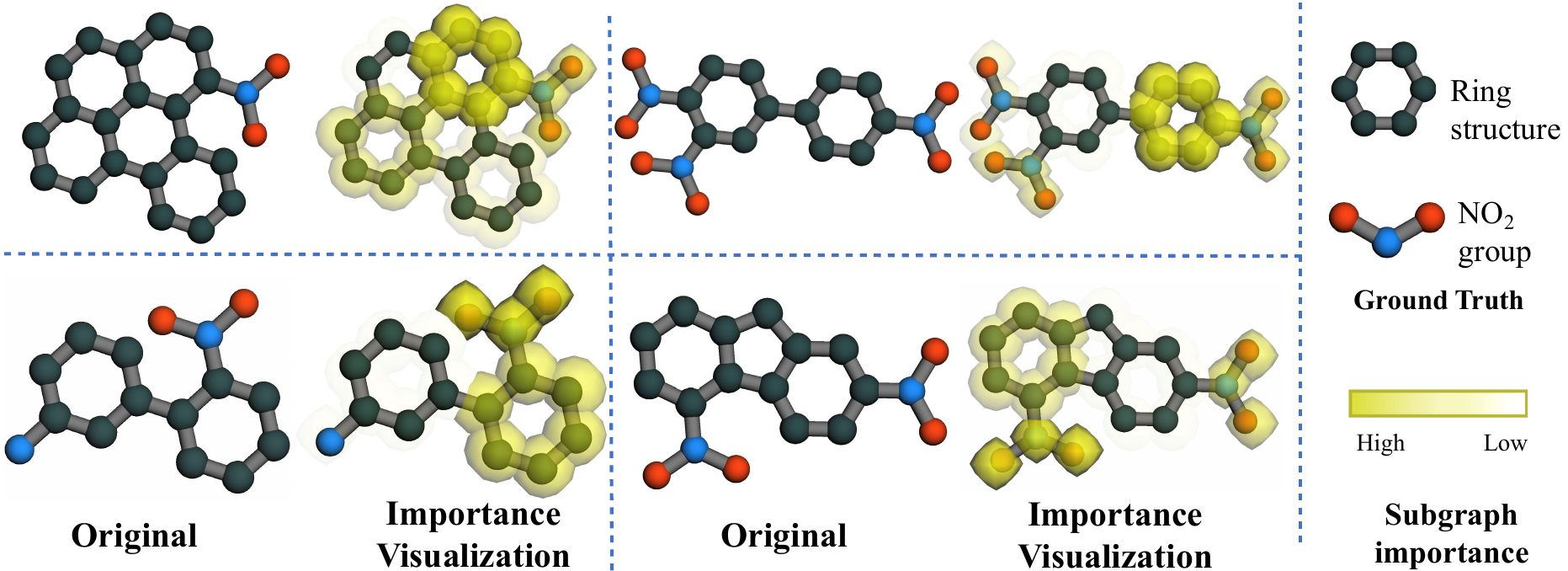} 
	\caption{We visualize the graphs in the MUTAG dataset, the deeper color indicates the more important subgraphs and the ground truth is ring structure and $\text{NO}_2$ group.} 
	\label{fig:subgraph}
\end{figure}

To validate the effectiveness of our \ours{} framework (see Fig.~\ref{fig:subgraph}), we trained a view generator on MUTAG dataset~\cite{debnath1991structure, DBLP:conf/icml/KriegeM12}, which is a real-world chemical dataset, we evaluate the effectiveness of \ours{} that identifies important partitioned subgraphs based on chemical domain knowledge. It is known that carbon rings and $\text{NO}_2$ groups tend to be mutagenic~\cite{debnath1991structure}, and we study whether the view generator identifies these important structures. The visualization results explain the capability of \ours{} as it has successfully identified the carbon rings and $\text{NO}_2$ groups as critical subgraphs.
The results emphasize the importance of capturing subgraph features and relationships among subgraphs, and prove the effectiveness of our \ours{}.

\subsection{Ablation Study}
The proposed method first partitions the graph into a set of subgraphs, and then jointly trains the subgraph augmentation selector and the subgraph view generator to generate graph views. In this subsection, we present an ablation study on the unsupervised learning task to evaluate the contribution of the three components as shown in Table~\ref{tab:ablation-exp}. We use "P" as a suffix of the model name (\ours{}-P) to denote the model without a subgraph partition module, a suffix "S" (\ours{}-S) to denote the model with a random selector for subgraph augmentation strategy rather than a learnable one, and a suffix "R" (\ours{}-R) to denote the model with a random subgraph view generator (e.g., random node dropping in one subgraph) rather than a learnable one. For variations of the subgraph partitioning algorithm, we evaluate \ours{} using the Louvain algorithm and Girvan-Newman (GN) algorithm respectively, and provide the average subgraph number of a partitioned graph in each dataset. The results indicate that the \ours{} framework performs well across different subgraph partitioning algorithms, highlighting its effectiveness and robustness. Furthermore, the extensive experiments across several datasets demonstrate the robustness and adaptability of \ours{}, indicating its performance is resilient to variations in the partitioned subgraph sizes.

\begin{figure}
	\centering \includegraphics[width=0.8\linewidth]{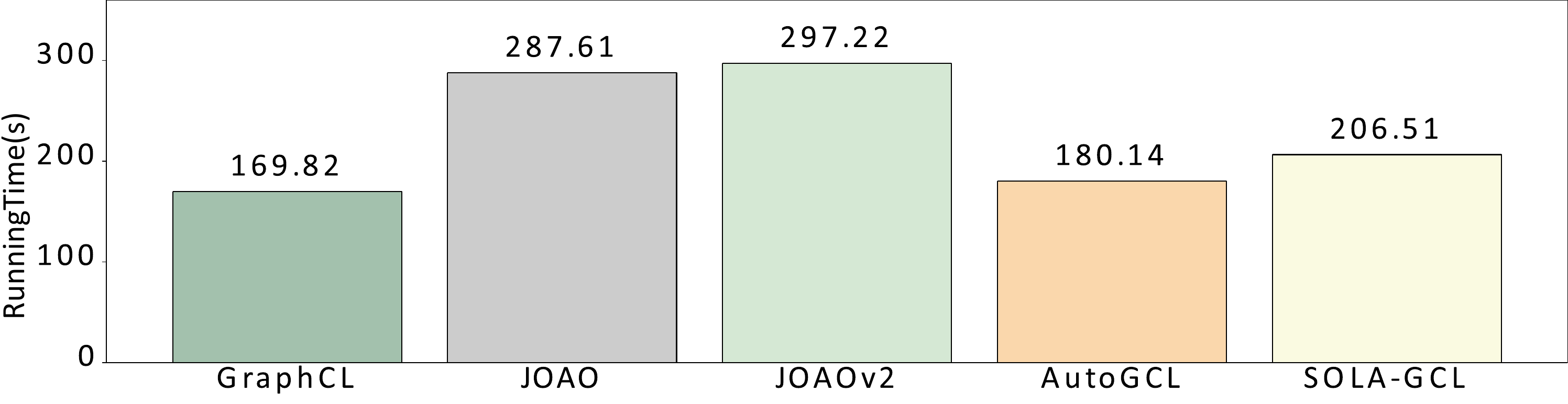} 
	\caption{Comparisons in terms of real running time. Each model is trained for 100 epochs on each dataset and the average training time of one epoch is reported.} 
	\label{fig:running_time}
\end{figure}

\subsection{Efficiency Analysis}
\label{app:convergence}
As shown in Fig.~\ref{fig:running_time}, we present the comparisons on training time consumption for each epoch in the training process. While GraphCL and AutoGCL show the lowest training time consumption, \ours{} strikes an optimal balance between efficiency and effectiveness. It achieves better performance than JOAO and JOAOv2 and consumes less training time, demonstrating a significant advantage in both performance and efficiency.

\section{Conclusion}

In this work, we present \ours{}, a novel data augmentation method for graph contrastive learning. In contrast to existing methods that overlook the importance of the critical role of subgraph structures, \ours{} takes a comprehensive approach by considering the information of intra-subgraph characteristics and inter-subgraph relationships. 
This is achieved through the joint training of a subgraph augmentation selector and a subgraph view generator, enabling the generation of learnable graph views. 
Across a range of settings and datasets, our data augmentation method consistently surpasses the performance of current approaches, underscoring its superiority.

\bibliography{aaai25}

\end{document}